\documentclass{article}

\usepackage{PRIMEarxiv}

\usepackage[utf8]{inputenc}
\usepackage[T1]{fontenc}
\usepackage{microtype}

\usepackage{amsmath,amssymb,amsfonts,mathtools}
\usepackage{mathrsfs}
\usepackage{stmaryrd}
\usepackage{amsthm}

\theoremstyle{definition}

\numberwithin{equation}{section}

\usepackage{graphicx}
\graphicspath{{./}}
\usepackage{float}
\usepackage{array}
\usepackage{multicol,multirow}
\usepackage{rotating}
\usepackage{pifont}
\usepackage[table]{xcolor}
\usepackage{algorithm}
\usepackage{algpseudocode}
\usepackage{appendix}
\usepackage{ifpdf}
\usepackage{lineno}
\usepackage{textcomp}
\usepackage{svg}

\usepackage{url}
\usepackage{natbib}
\usepackage{hyperref}

\title{A Scale-Adaptive Framework for Joint Spatiotemporal Super-Resolution with Diffusion Models}

\author{
  Max Defez$^{1,2}$,
  Filippo Quarenghi$^{1,2}$,
  Mathieu Vrac$^{3}$,
  Stephan Mandt$^{4}$ and
  Tom Beucler$^{1,2}$\\[0.5em]
  $^{1}$Faculty of Geosciences and Environment, University of Lausanne, Lausanne, VD, Switzerland\\
  $^{2}$Expertise Center for Climate Extremes, University of Lausanne, Lausanne, VD, Switzerland\\
  $^{3}$Laboratory for Climate and Environmental Sciences, Gif-sur-Yvette, France\\
  $^{4}$Department of Computer Science, University of California, Irvine, California, United States of America\\
}

\begin{document}
\maketitle

\begin{abstract}
Deep-learning video super-resolution has progressed rapidly, but climate applications typically super-resolve either space or time, and joint spatiotemporal models are often designed for a single pair of super-resolution (SR) factors, limiting transfer across spatial resolutions and temporal cadences (frame rates). We present a scale-adaptive framework that reuses the same architecture across factors by decomposing spatiotemporal SR into a deterministic prediction of the conditional mean, with attention, and a residual conditional diffusion model, with an optional mass-conservation transform to preserve aggregated totals. Assuming that larger SR factors primarily increase underdetermination (hence required context and residual uncertainty) rather than changing the conditional-mean structure, scale adaptivity is achieved by retuning two factor-dependent hyperparameters before retraining: the diffusion noise schedule amplitude $\beta$ (larger for larger factors to increase diversity) and the temporal context length $L$ (set to maintain comparable attention horizons across cadences), and optionally a third, the mass-conservation function $f$ (tapered to limit the amplification of extremes for large factors).  Demonstrated on reanalysis precipitation over France (Coméphore), the same architecture spans super-resolution factors from 1 to 25 in space and 1 to 6 in time, yielding a reusable architecture and tuning recipe for joint spatiotemporal super-resolution across scales.
\end{abstract}


\section*{Plain Language Summary}
We propose a ready-to-use diffusion-based video super-resolution architecture that reuses the same model design across spatial and temporal refinement settings, requiring only 2-3 retuned hyperparameters before retraining. This enables reuse across environmental datasets without redesigning the architecture. Demonstrated on precipitation, a challenging zero-inflated, heavy-tailed variable, the model generates realistic ensembles across scales, supporting climate-impact applications such as downscaling.

\section{Introduction: Spatiotemporal Super-Resolution as Conditional Generative Modeling}
Video Super-Resolution (VSR) aims to recover high-fidelity, high-resolution (HR) frame sequences from their low-resolution (LR) counterparts by exploiting spatiotemporal correlations to resolve details. While data-driven methods, namely deep learning algorithms ranging from deterministic regression models (e.g., transformers, recurrent neural networks) to generative models (e.g., diffusion models), have dominated the state-of-the-art VSR in computer vision, the adoption of data-driven VSR for climate science remains rare (\cite{Liu2022}, \cite{zhang2025deeplearningempoweredsuperresolution}, \cite{GOPALAKRISHNAN2024117175}). 

A central barrier is scale adaptivity. Most deep-learning VSR models are designed and trained for a single super-resolution setting (e.g., $\times 4$) and a fixed combination of spatial grid and temporal cadence. This rigidity is particularly limiting in atmospheric science, where products from satellites, radar mosaics, gauges, and reanalyses come with heterogeneous resolutions and sampling frequencies. As a result, practitioners often fully redesign model architectures for each new dataset or target resolution, making pipelines problem-specific and computationally costly.

We address this gap in architecture design by focusing on the spatiotemporal super-resolution of precipitation data. Precipitation is a particularly challenging testbed due to its high intermittency, non-Gaussian statistics, and complex multiscale interactions (\cite{https://doi.org/10.1029/JD092iD08p09693}). Accurate spatiotemporal VSR in this context is not merely about aesthetic enhancement; it is essential for resolving localized extreme events and temporal interactions for impact models (e.g., flash flood forecasting) as demonstrated by \cite{OCHOARODRIGUEZ2015389} and \cite{hess-21-3859-2017}. 

Research in precipitation nowcasting provided foundational VSR architectures to model spatiotemporal atmospheric dynamics. Early deep learning approaches, such as the ConvLSTM network by \cite{shi2015convolutionallstmnetworkmachine}, demonstrated that neural networks modeling sequences could effectively capture the temporal evolution of rain fields. 

However, these early deterministic methods share a fundamental limitation: by optimizing errors at the pixel level (e.g., Mean Squared Error), they smooth out high-frequency details and lead to a systematic underestimation of extremes.

To address this spectral bias, precipitation nowcasting shifted toward generative models capable of modeling the stochasticity inherent in fine-scale weather systems. \cite{Leinonen_2021} introduced a foundational recurrent conditional Generative Adversarial Network (cGAN) specifically designed to super-resolve time-evolving atmospheric fields. Building on this, \cite{glawion2025global} recently demonstrated the utility of cGANs for generating consistently evolving global rain fields. However, despite their ability to recover high-frequency textures, GANs remain prone to training instabilities and mode collapse, often failing to capture the full probability distribution of the data \cite{tomasi2025LDM_COSMO}.

Recent progress therefore increasingly relies on probabilistic diffusion models. This transition is again paralleled in the nowcasting literature, where \cite{yu2024diffcastunifiedframeworkresidual} introduced DiffCast to separate deterministic motion trends from stochastic residuals. This decomposition, separating bulk advection from fine-scale texture, is equally critical for spatiotemporal super-resolution, as demonstrated by the success of the SpatioTemporal Video Diffusion (STVD) architecture from \cite{stvd}.

Despite these advances, diffusion-based VSR architectures, including STVD, are typically tuned for fixed grids and cadences, limiting transfer across SR factors $(S,T)$. Our contribution is a modeling principle for architectural reuse: we decompose joint spatiotemporal SR into (i) a deterministic predictor of the conditional mean and (ii) a diffusion model of stochastic residuals. We assume that increasing SR factors mainly increases underdetermination (and thus the required temporal context and residual uncertainty) without fundamentally changing the conditional-mean structure. Accordingly, for each $(S,T)$ we train a separate instance with shared architecture and training procedure (but not shared weights), and retune only two factor-dependent hyperparameters (attention context length and diffusion noise schedule) to calibrate context and residual variance, rather than redesigning the architecture. 

\section{Methodology}

\subsection{Problem Definition and Notation}
To explore the design space of spatiotemporal SR factors, we consider pairs $(S,T)\in{\mathbb{N}^*}^2$, where $S$ and $T$ denote the spatial and temporal SR factors, respectively, and use this sweep to guide hyperparameter choices. Let $H$ and $W$ denote the HR frame height and width, in our case $H = W = 100$. An HR observation is a nonnegative frame in $\mathbb{R}_+^{H\times W}$. We index frames by time $t\in\mathcal{T}$ and tile $l\in\mathcal{L}$, and denote the corresponding HR frame by $y^{t,l}\in\mathbb{R}_+^{H\times W}$.

In our experiments, we adopt a ``perfect-model setting'' in which LR frames are a deterministic coarsening of HR frames, computed by averaging over non-overlapping $T$ temporal blocks and $S\times S$ spatial blocks (see Appendix~\ref{app:block_avg} for the formal averaging operator). The resulting LR frame is denoted $x^{t,l}\in\mathbb{R}_+^{(H/S)\times(W/S)}$. Leveraging temporal autocorrelation, the input is a LR sequence of length $L$ ending at $t$, i.e.,
$x^{t,l}(L)\equiv\big(x^{t-(L-1)T,l},\,\ldots,\,x^{t,l}\big)\in\big(\mathbb{R}_+^{(H/S)\times(W/S)}\big)^L$.
We also condition on a static auxiliary field $M_l$ (tile topography), it is the only additional input used in the model.

Given an LR sequence ending at time $t$, our model predicts the corresponding HR block without future information:
$\big(\hat y^{t,l},\ldots,\hat y^{t+T-1,l}\big)=\Phi_{S,T}\!\left[x^{t,l}(L),M_l\right]$,
where $\Phi_{S,T}$ is the SR mapping for factors $(S,T)$; our goal is to develop a versatile architecture that can learn $\Phi_{S,T}$ for any $(S,T)$. The overall architecture is identical for all $(S,T)$, except the 2-3 hyperparameters described in Section~\ref{sub:HP}. However, each $(S,T)$ has its own model instance with specific weights learned from data at the corresponding resolution.

\subsection{A Residual Diffusion Architecture for Scale-Adaptive Super-Resolution}

To motivate a generative approach, consider $(S,T)=(25,6)$: the model must reconstruct $T=6$ HR frames of size $H\times W=100\times100$ from a single LR frame of size $(H/S)\times(W/S)=4\times4$, i.e., infer $6\times25^2=3{,}750$ HR pixels per LR pixel. This mapping is highly non-injective, so we model a conditional distribution rather than a single point estimate, allowing for multiple plausible scenarios. To stabilize learning, we sample stochastic outputs as residual refinements around a deterministic mean prediction, rather than generating videos from scratch. Accordingly, we use a two-stage architecture, loosely inspired by \cite{stvd}: a deterministic U-Net predicts a mean HR block, which conditions a diffusion model that samples residual videos. Both components include spatiotemporal attention and an optional mass-conservation transform to preserve LR-aggregated totals.

\begin{figure}[hbtp]
    \centering
    \includegraphics[width=0.95\linewidth]{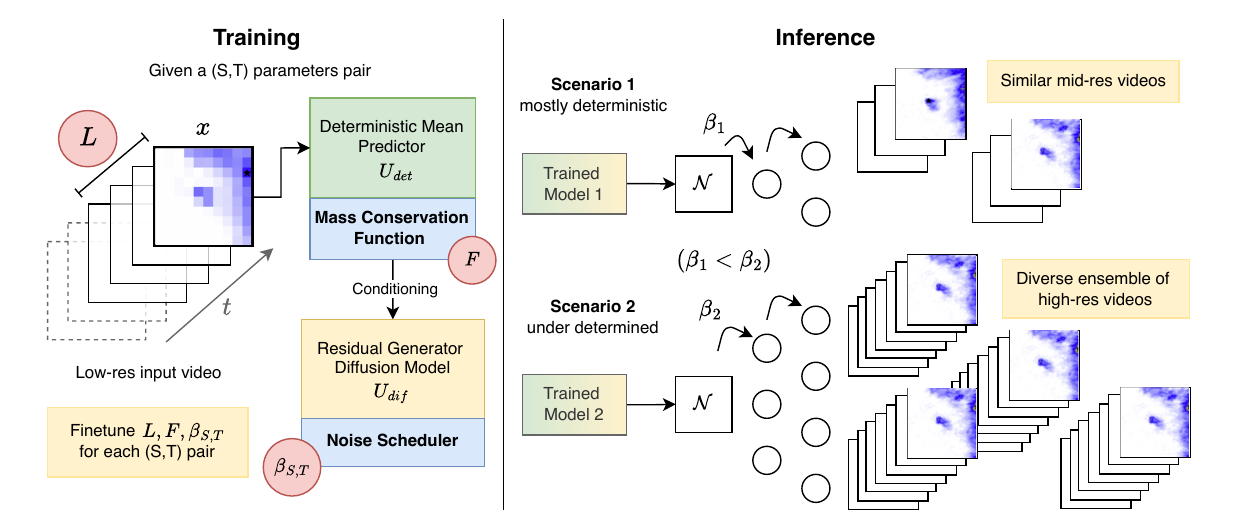}
    \caption{Scale-adaptive spatiotemporal VSR: a deterministic U-Net $U_{\mathrm{det}}$ predicts a coarse mean field, followed by an optional mass-conservation transform and a residual diffusion head $U_{\mathrm{dif}}$ that generates target-resolution videos conditioned on the mean. Scale adaptivity is achieved by tuning the attention horizon $L$ and the noise schedule $\beta$ (increased for larger SR factors to promote video diversity)}
    \label{fig:placeholder}
\end{figure}

\subsubsection{Deterministic Mean Predictor: U-Net with Spatiotemporal Attention and Mass Conservation}
For each $(S,T)$, we train a deterministic U-Net $U_{\mathrm{det}}$ to predict a mean HR sequence $D^{t,l}\in(\mathbb{R}_+^{H\times W})^T$ from the LR context $x^{t,l}(L)$ and static topography $M_l$. Each LR frame in $x^{t,l}(L)$ is bicubically interpolated to an $H\times W$ frame, yielding $\mathrm{BI}(x^{t,l}(L))$; concatenating these $L$ frames with $M_l$ as an extra channel gives an $(L{+}1)$-channel input, and the network outputs $T$ channels at the same resolution.

Starting from the U-Net of \cite{ronneberger2015unetconvolutionalnetworksbiomedical}, we augment each encoder stage with factorized self-attention. Temporal attention operates at fixed spatial locations: for each $(i,j)$, we compute multi-head self-attention over $\big(x^{t-(L-1)T,l}_{i,j},\ldots,x^{t,l}_{i,j}\big)$ like \cite{vaswani2023attentionneed}. We also use windowed spatial self-attention within each frame, where attention is calculated over a $(2n{+}1)\times(2n{+}1)$ neighborhood for each location, with window sizes decreasing with depth. With 4 encoder stages and the bottleneck, we set the window size tuple to $N=(n_1,n_2,n_3,n_4,n_b)=(3,3,1,1,1)$ and use 4 heads for both temporal and spatial attention. Optionally, we enforce mass conservation between the LR input and the predicted HR sequence via frame-level rescaling introduced by \cite{harder2024hardconstraineddeeplearningclimate}. Let $\tilde D^{t,l}=F\!\left(U_{\mathrm{det}}([\mathrm{BI}(x^{t,l}(L)),M_l])\right)$ be a nonnegative intermediate prediction obtained by applying a nonnegative function $F$ element-wise. We scale the $T$ predicted HR frames so that their total mass matches the LR aggregate:

\begin{equation}
(D^{t+k,l})_{i,j}
=
(\tilde D^{t+k,l})_{i,j}\,
\frac{S^2T\sum_{p=0}^{H/S-1}\sum_{q=0}^{W/S-1} x^{t,l}_{p,q}}
{\sum_{k'=0}^{T-1}\sum_{i'=0}^{H-1}\sum_{j'=0}^{W-1}(\tilde D^{t+k',l})_{i',j'}},
\qquad k=0,\ldots,T-1,
\label{mass_conservation}
\end{equation}

where the numerator is the LR-implied total mass over the corresponding $T$-frame HR sequence (Appendix~\ref{app:block_avg}). We denote this optional step by $D^{t,l}=\mathrm{MC}\!\left(F\!\left(U_{\mathrm{det}}([\mathrm{BI}(x^{t,l}(L)),M_l])\right),\,x^{t,l}\right)$; the choice of $F$ is treated as a factor-dependent hyperparameter (Section~\ref{sub:Mass_Conservation_fx}). See Appendix \ref{sub:U_Net} for architectural details. For this first deterministic step, our scale-adaptivity principle is to use a shared architecture across SR factors except for the attention window length $L$ and mass-conservation function $F$, while allowing for different weights.

\subsubsection{Stochastic Residuals: Scale-Dependent Uncertainty via Residual Diffusion}
Given the deterministic mean prediction $D^{t,l}$, we model the remaining fine-scale uncertainty with a conditional diffusion model of residuals $r^{t,l}\equiv y^{t,l}-D^{t,l}$, the final model prediction is then $\hat y^{t,l}=D^{t,l}+\hat r^{t,l}$. Because the final prediction is $\hat y = D + \hat r$, the ensemble mean equals $D$ plus the mean of the sampled residuals, so it inherits the bias of $D$; each member adds its own spread on top, so a single member always has a higher MSE than the ensemble mean. This is why bias metrics are reported on the ensemble mean as well as per member (Section~\ref{sub:evaluation}). The scale-adaptivity principle in this second step is architectural reuse, we keep the same residual-diffusion architecture across SR factors and express increased underdetermination mainly as increased residual uncertainty by calibrating the noise schedule $(\beta_j)_j$, rather than through fundamental architectural changes. This relies on the assumptions that (i) the conditional-mean structure learned by $U_{\mathrm{det}}$ is relatively stable across $(S,T)$, (ii) changing $(S,T)$ primarily affects the amplitude of residual uncertainty (spread) more than its qualitative structure, and (iii) task-level calibration via $(\beta_j)_j$ is sufficient. For $(S,T)=(10,3)$, assumption (iii) holds for the distribution of generated values but not pixel by pixel: the ensemble selected this way is under-dispersed (Figure~\ref{fig:calibration}).

We implement the residual generator as a denoising diffusion probabilistic model from \cite{ho2020denoisingdiffusionprobabilisticmodels} with $J=1000$ steps. During training, we sample $j\sim\mathcal{U}\{1,\ldots,J\}$, add Gaussian noise to the true residual $r^{t,l}$ according to $\beta_j$ (Appendix~\ref{forward_pass}), and train a diffusion U-Net $U_{\mathrm{dif}}$ to predict the ``velocity'' (a weighted sum of the noise and the original data, see Appendix~\ref{velocity}) using an $\ell_2$ loss \cite{salimans2022progressivedistillationfastsampling}. The diffusion U-Net takes as input the deterministic mean $D^{t,l}$, the bicubically interpolated frame to downscale $\text{BI}(x^{t, l})$, the noised residual, and a learned 128-dimensional embedding of $j$. It uses the same windowed spatial self-attention as the deterministic U-Net but also adds cross-attention on the whole bicubic interpolation's output $\text{BI}(x^{t, l}(L))$. The context for this cross-attention is produced by a small temporal encoder that, in our implementation, is randomly initialised and not trained. The trained diffusion U-Net turned out to be insensitive to it: replacing the context by zeros changes its output by less than $10^{-6}$ in relative terms. In practice the diffusion head is therefore conditioned on $D$ and on the last bicubic frame, and information from earlier input frames reaches it only through $D$.

At inference time, we draw $\hat r^{t,l}(J)\sim\mathcal{N}(0,I)$ and apply the learned reverse process for $j=J,\ldots,1$ with the same conditioning to obtain $\hat r^{t,l}(0)$, then reconstruct $\hat y^{t,l}=D^{t,l}+\hat r^{t,l}(0)$ (see Appendix~\ref{sub_reverse_process_diffusion}). If enabled, we apply the same mass-conservation step as in Eq.~\eqref{mass_conservation} using the same $F$. Training and sampling pseudocode is given in Appendix~\ref{algorithm_forward_reverse}, and the overall architecture is shown in Appendix~\ref{sub:overall_architecture}.

\subsection{Tunable Hyperparameters for Scale Adaptivity\label{sub:HP}}

In this section we present the three tunable hyperparameters that enable the model to maintain a form of relevant scale adaptivity: respectively the length of input $L$, the variance scheduler $\beta$ and the mass-conservation function $F$. Each hyperparameter will be optimized for each pair of SR factors.

\subsubsection{Length $L$ and Attention Time $A_T$}

$L$ is defined as the number of consecutive frames, including the one to downscale to give the model. We introduce the attention time $A_T \equiv T\times L$ which is easier to interpret as it represents the temporal horizon, expressed in hours, up to which the model retains information about past data. However, one will optimize the hyper-parameter $L$ for the sake of simplicity, which is clearly equivalent.

\subsubsection{Mass Conservation Function $F$\label{sub:Mass_Conservation_fx}}
\label{def_mc_function}

$F$ is the function that will be applied to the outputs of the deterministic and diffusion U-Nets. More specifically, our final prediction is the output of this function which is why $F$ should be positive, increasing and continuous in order to satisfy physical constraints. 

Given the mass-conservation equation \ref{mass_conservation}, considering a rapidly increasing function would be equivalent to applying a high-pass filter to the frames, which would lead to an overrepresentation of high precipitation values. Conversely, considering a flat function would be akin to blurring the image by predicting mid precipitation values everywhere. Thus, $F$ serves as a tool for handling model calibration, which can be assessed using the Probability Integral Transform Deviation (PITD), see Appendix \ref{PITD}. 

\subsubsection{Variance Scheduler $\beta$}

For the sake of simplicity, the variance scheduler defined by the sequence $(\beta_j)_{j\in [1, J]}$ will be characterized by the tuple $(\beta_{min}, \beta_{max})$, such that $ \forall j, \: \beta_j = \beta_{min} + \frac{j}{J}(\beta_{max} - \beta_{min})$. 

We fix $\beta_{min}$ to a small, standard value ($10^{-4}$ as suggested by \cite{ho2020denoisingdiffusionprobabilisticmodels}) to ensure a gentle corruption of clean inputs, and focus the optimization on $\beta_{max}$, which primarily controls the overall noise amount and has a stronger influence on generation quality.

\section{Application: Super-Resolution of Observed Precipitation across Spatiotemporal Factors}

The aim of this paper is to assess the versatility of this architecture by evaluating it across various super-resolution factors, while allowing only the parameters describe in section \ref{sub:HP} to change. Each pair of super-resolution factor corresponds to a distinct training instance and therefore to different learned weights. To evaluate the model performance, the following factor pairs $(S, T)$ were used: $(1, 3)$, $(10, 1)$, $(10, 3)$ and $(25, 6)$.

The observations were taken from the MétéoFrance dataset Coméphore, a precipitation reanalysis produced by merging radar and rain gauge observations, covering metropolitan France. Its spatial and temporal resolutions are 1 km and one hour. From a spatial perspective, the dataset covers a 1536-by-1536 rectangle representing the area whose northwestern corner is located at (53.670 N, -9.965 E).

The training and testing dataset are made of the entire 2023 and 2024 data, respectively. For both datasets, the analysis is restricted to the 400-by-400 square region with its northwestern corner at (49.400 N, -0.971 E). This area was sliced in 4$\times$4 square tiles of shape H$\times$W used for 4-fold cross-validation during training. As shown in Appendix \ref{cross_val_slicing}, each fold included exactly one tile per row and per column without overlapping to avoid topography overfitting.

In this dataset, precipitation could rise to extremely large values such as $200$ millimeter per hour. Given these unexpectedly high values considering the location, we decided to deal with outliers. Deleting them was not convenient because it also meant deleting the $L-1$ samples before and after the outlier, which would have significantly bias the model and decrease the length of the datasets. Thus the outliers were capped at very high but plausible values. \cite{WhyDoPrecipitationIntensitiesTendtoFollowGammaDistributions} showed precipitation tends to follow a gamma distribution. After fitting this distribution to our training dataset, the outliers were set to the $99.5^{th}$ percentile which amounted to $55$ millimeter per hours.

Given that the data are positive, it was especially relevant to apply min-max normalization to precipitation, with the maximum value computed from the training set being $55$ because of the previous step.

Finally, topography was incorporated into the model input so that it can capture potentially complex relationships, given that links between these two variables have been extensively documented, notably by \cite{StatisticalRelationshipsbetweenTopographyandPrecipitationPatterns}. The topographic data are from the GLO-90 Digital Surface Model (DSM) provided by the Copernicus Data Space Ecosystem. Its resolution is initially 90$\times$90 squared meter and was downsampled by average pooling to 1$\times$1 squared kilometer to make it homogeneous with the (bicubically interpolated) inputs. Topography data were positive and did not include obvious outliers, it was min-max scaled to $[0, 1]$ as well.

\section{Training Protocol, Hyperparameter Optimization, and Evaluation}

The training was managed with Adam optimizer from \cite{kingma2017adammethodstochasticoptimization}, with an initial learning rate of $10^{-4}$ decreasing to $0$ over $80$ epochs (where no significant progress was observed) through a cosine annealing, defined by \cite{loshchilov2017sgdrstochasticgradientdescent}. Early stopping is applied on the validation loss (the training loss evaluated on held-out tiles), which is computed every $3$ epochs; training stops after $8$ evaluations without improvement. In practice this ended training well before the maximum number of epochs. Due to the high memory cost of spatial and temporal attention, the model is constrained to batch sizes of $12$ for our HPC cluster.

\subsection{Hyperparameter Optimization}

We explain the method used to determine the optimal hyperparameters of interest $(L, F, \beta)$ for each pair of factors.

\textit{Length $L$}: We select the temporal length $L$ using an elbow-based procedure. Starting from small values, $L$ is progressively increased and the corresponding validation performance is monitored. The chosen value corresponds to the smallest $L$ beyond which performance gains become marginal, ensuring a trade-off between accuracy and computational cost.

\textit{Mass Conservation Function $F$}: We restrict $F$ to power functions and introduce a thresholded ReLU $r_\alpha(x) = max(0, x - \alpha)$, this ReLU is applied before and after any mass-conservation function $F$ to suppress negligible precipitation values and keep only the significant and realistic rain. The mass-conservation transform is applied only after the $20^{th}$ training epoch to avoid early training instabilities. The exponent of $F$ is tuned using the PITD: if the model is over-dispersive, the growth rate of F is reduced, and conversely. In cases where the ReLU leads to a zero denominator in Eq. \ref{mass_conservation}, only the ReLU is applied.

\textit{Variance Scheduler $\beta$}: Considering the modeling choices of $\beta_j$, only the value of $\beta_{max}$ is to be determined. Increasing $\beta_{max}$ corresponds to injecting more noise into the image and thus increasing the variability in the sampling of scenarios. The value of $\beta_{max}$ is also determined with the PITD such that the entire set of scenarios remains well calibrated and plausible. Because the PITD compares distributions of values within frames (Appendix~\ref{description_metrics}), this criterion makes the generated fields statistically realistic but does not by itself make the ensemble calibrated pixel by pixel; for $(S,T)=(10,3)$ the selected value leaves the ensemble under-dispersed (Figure~\ref{fig:calibration}). Indeed, a small value of $\beta_{max}$ will typically lead to a blurred video close to the deterministic prediction, missing to represent high frequencies and extreme values whereas too large values introduce unrealistic high-frequency artifacts.

\textit{Computing time}: All the models were run on a NVIDIA A100-PCIE-40GB. For a specific pair $(S, T)$, training a model takes one day, while inference time is proportional to the number of generated scenarios: with $1000$ denoising steps, generating one member for a batch of $12$ samples takes about $1.5$~min, so evaluating $8$ members over the full test set of $(10,3)$ ($39{,}104$ samples) takes about $650$ GPU-hours. A significant avenue for improving this work could be to find ways to perform inference more quickly, for example by using architectures from \cite{song2022denoisingdiffusionimplicitmodels}, \cite{kong2021fastsamplingdiffusionprobabilistic} or \cite{rombach2022highresolutionimagesynthesislatent}.

\textit{Practical deployment}: Applying the framework to a new target proceeds as follows. Given a desired spatial and temporal super-resolution pair $(S,T)$, the practitioner reuses the architecture unchanged and retunes only the factor-dependent hyperparameters --- the input length $L$, the noise-schedule amplitude $\beta_{\max}$, and the mass-conservation function $F$ --- following the PITD-based recipe of Section~\ref{sub:HP} (with the caveats of Section~\ref{sub:best_hp}), then trains a single instance for that pair. For a target pair close to one already trained, retraining can be avoided: the low-resolution field is first brought to the nearest learned resolution by a light interpolation and then passed through the corresponding trained model. Inference is the main cost: with $1000$ denoising steps, one member for a batch of $12$ samples takes about $1.5$~min on a single A100, so the method suits offline analysis and targeted scientific studies rather than real-time operational use.

\subsection{Evaluation Metrics\label{sub:evaluation}}

We compared five models, three of which were developed in the present article. Two baselines were employed in this study: bicubic and nearest neighbor interpolation. As for our three models, we considered the full architecture, a deterministic-only variant (without diffusion), and a variant in which both temporal and spatial attention mechanisms were ablated.
All five models were compared based on $8$ climate and vision metrics, detailed in Appendix \ref{description_metrics}. For the generative models, the bias metrics (MSE, MAE and 99th~PE) are computed from the \emph{ensemble mean} of the generated members, with the mean per-member score reported alongside; the distributional metrics (CRPS, PITD) are computed over the pooled members. The deterministic models apply the threshold ReLU and the mass-conservation function $F$ to their output; the full architecture applies them to the deterministic mean and again to every member; the interpolation baselines are not post-processed. The full per-metric protocol is given in Appendix~\ref{description_metrics}.

\section{Results}

\subsection{Example of Video Super-Resolution}
\label{example_video}

\begin{figure}[hbtp]
  \centering
  \includegraphics[width=\textwidth]{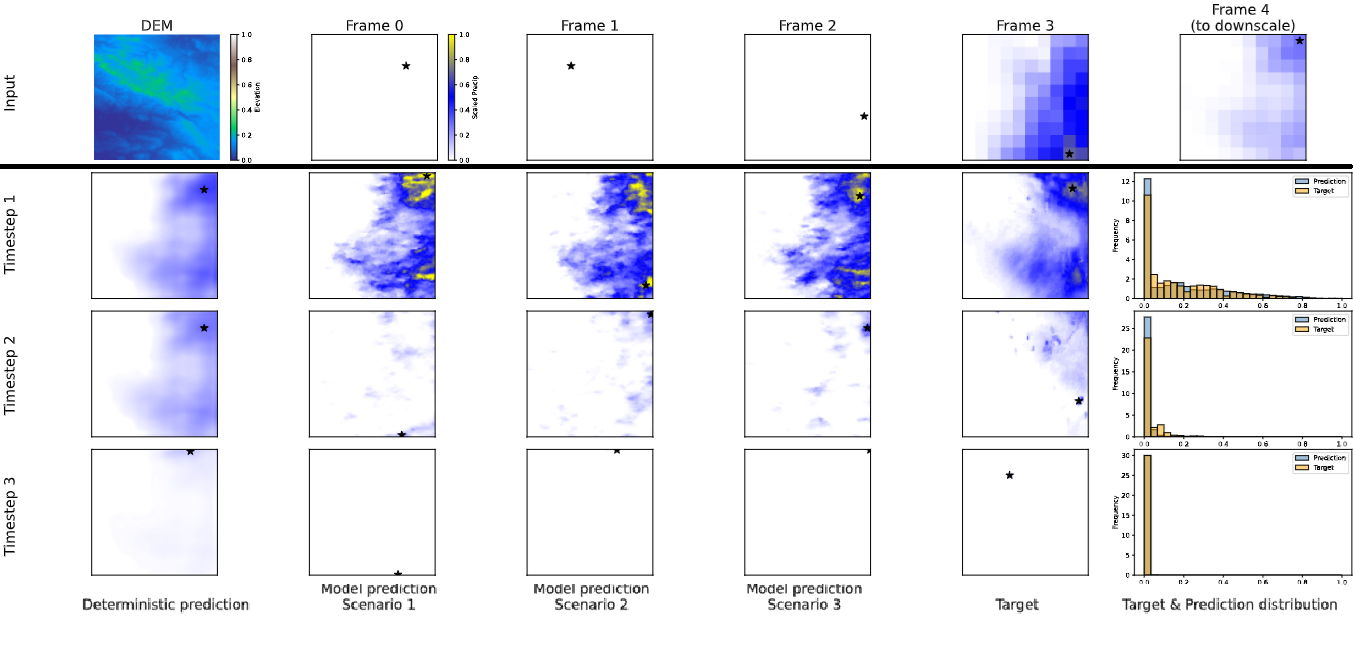}
  \caption{Qualitative example of spatiotemporal precipitation super-resolution produced by the proposed model. The first row shows the low-resolution input sequence and the topography. The first column corresponds to the deterministic (average) prediction, while each subsequent column displays a high-resolution scenario generated by the diffusion model, illustrating the diversity of plausible outcomes}
  \label{video_prediction}
\end{figure}

Figure~\ref{video_prediction} presents an example of video downscaling with super-resolution factors $(10, 3)$. The first row shows the model inputs, namely the topography and the sequence of low-resolution frames from past to present. In this example $L$ is set to 5 so the model takes 5 frames as input.

The first column displays the output of the deterministic U-Net, which is inherently smooth and blurry, as it represents the mean prediction. The next three columns correspond to three distinct generated scenarios. Each row represents a timestep, and since the temporal super-resolution factor is set to 3, three rows are shown. The super-resolved frames should be compared to the ground truth provided in the fifth column. Although the scenarios differ from one another, they consistently capture the large-scale evolution of precipitation, correctly indicating a decrease in rainfall at later timesteps which makes the predictions very satisfactory. In particular, the extreme values shown in yellow are well reconstructed by the model, which is naturally not the case for the deterministic model and the two baselines.

\subsection{Quantitative Benchmarking}

The quantitative comparison between models is reported in Table 1. It assess both reconstruction accuracy and the preservation of precipitation structure and extremes.

\begin{table}[hbpt]
\centering
\small
\begin{tabular}{lcccccccc}
\hline
\rowcolor{lightgray!50} Architecture & MSE $\downarrow$ & MAE $\downarrow$ & 99th PE $\downarrow$ & LSD $\downarrow$ & EMD $\downarrow$ & SSIM $\uparrow$ & PITD $\downarrow$ & CRPS $\downarrow$ \\
\hline
Bicubic interp.
 & 2.97E-3 
 & 1.47E-2 
 & 6.11E-2 
 & 3.12
 & \cellcolor{green!25}1.06E-3 
 & 8.70E-1 
 & 1.12E-1 
 & 1.47E-2 \\ \hline
Nearest Neighbor 
 & 3.05E-3 
 & 1.48E-2 
 & 6.19E-2 
 & 3.27
 & 1.12E-3
 & 7.90E-1 
 & 1.09E-1
 & 1.48E-2 \\ \hline

\cellcolor{blue!25}Deterministic 
 & \cellcolor{green!25}2.81E-3 & \cellcolor{green!25}1.35E-2 & \cellcolor{green!25}5.37E-2 & \cellcolor{green!25}2.76 & 1.12E-3 & \cellcolor{green!25}8.81E-1 & 1.65E-1 & 1.35E-2  \\ \hline

\cellcolor{blue!25}Generative \\\cellcolor{blue!25}(no attention) 
 &  5.11E-3 & 1.76E-2 & 5.61E-2 & 7.37 & 5.49E-3 & 8.60E-1 & 5.60E-2 & \cellcolor{green!25}1.22E-2 \\ \hline
\cellcolor{blue!25}Full architecture \\\cellcolor{blue!25}(ensemble mean)
 & 3.93E-3
 & 1.58E-2
 & 5.73E-2
 & 3.22
 & 4.95E-3
 & 8.73E-1
 & \cellcolor{green!25}1.60E-2 & \cellcolor{green!25}1.22E-2 \\ \hline
\cellcolor{blue!25}Full architecture \\\cellcolor{blue!25}(per member)
 & 5.30E-3
 & 1.81E-2
 & 6.76E-2
 & 2.96
 & 5.65E-3
 & 8.57E-1
 & -- & -- \\ 
\hline
\end{tabular}
\caption{Performance comparison for the pair $(10, 3)$ between two baseline approaches and three proposed super-resolution models in blue. Models are assessed using eight complementary metrics capturing pixel-level accuracy, structural consistency, and climatological relevance. Best scores per metric are highlighted in green. For the full architecture, bias and image metrics are computed on the \emph{ensemble mean} of $8$ members and, separately, averaged over single members; CRPS and PITD use all members (Appendix~\ref{app:metric-computation}). These two rows are computed over the complete test set. The ``Generative (no attention)'' row comes from the original evaluation code: its bias and image metrics score a single member, so it compares with the per-member row, and its CRPS and PITD use $5$ members. CRPS depends on ensemble size (this estimator is higher for smaller ensembles), so CRPS values of ensembles of different size are not strictly comparable.}
\label{tab:results_10_3}
\end{table}

We interpret these results through the bias--variance decomposition. Since the deterministic model approximates the conditional mean, it minimises the MSE; a single generative member adds its own spread to the error, and the ensemble mean removes most of that spread but not all of it. Table~\ref{tab:results_10_3} shows exactly this ordering, $\text{MSE}_{\mathrm{det}} < \text{MSE}_{\mathrm{ens.\,mean}} < \text{MSE}_{\mathrm{member}}$, and we do not claim that the generative model improves pixel-wise accuracy. The deterministic model is best on MSE, MAE, 99th~PE, LSD and SSIM, and bicubic interpolation on EMD. The 99th~PE needs care: it compares, frame by frame, the 99th percentile of one prediction with that of the target, so it penalises a single member whose peak intensity varies from draw to draw, even when the members are right on average. On a subset of $192$ test samples, the deterministic model underestimates the 99th percentile of heavy-rain frames by $34\%$ on average, while the generated members underestimate it by $9\%$; the deterministic model still obtains the lower 99th~PE because its error varies less from frame to frame. Over all frames, the members also place some rain in $31\%$ of dry frames, against $21\%$ for the deterministic model. The ensemble mean is worse than the deterministic mean $D$ of the full model (MSE $2.84\times10^{-3}$) by more than its finite size explains ($8$ members leave about $0.2\times10^{-3}$ of sampling noise in the MSE, while the gap is $1.1\times10^{-3}$), so the mean of the generated members is slightly shifted away from $D$. The generative model's advantage lies in the probabilistic and spectral diagnostics: its ensemble has the lowest CRPS ($1.22\times10^{-2}$, against $1.35\times10^{-2}$ for the deterministic model, whose CRPS equals its MAE), its members have the most realistic distribution of rain values (lowest PITD), and they match the target power spectrum at all scales, whereas the deterministic prediction keeps only $15$--$20\%$ of the small-scale power (Figure~\ref{fig:rapsd}). The deterministic model's lower LSD comes from dry frames, where the LSD heavily penalises any predicted rain; in frames with rain, the generated members have the lower LSD (Appendix~\ref{app:metric-computation}). The ensemble is, however, under-dispersed pixel by pixel: the observation falls outside the range of the $8$ members in $33\%$ of the pixels instead of $22\%$, and the spread is about $40\%$ smaller than the error of the ensemble mean (Figure~\ref{fig:calibration}). We therefore present the generative head as a way to obtain an ensemble of realistic fields at a modest cost in pixel-wise accuracy, not as a uniformly better model nor yet as a calibrated one. Compared with the generative model without attention, the full architecture has a similar CRPS but a PITD lower by a factor of about $3.5$.
More specifically, we notice that the baselines perform worse on LSD and 99th PE, which supports our initial motivation that simple and coarse models struggle to capture the key features of precipitation data: climatic extremes (99th PE) and local-scale variations represented by high spatial frequencies (LSD). 

These results demonstrate that our final model is effective and versatile for tasks involving downscaling. Nevertheless, the diffusion module significantly increases inference time, and slightly increases training time. Given the decent performance of the purely deterministic model, one could reasonably rely on it for applications requiring rapid and computationally cheap predictions, or when pixel-wise accuracy rather than a realistic ensemble is the priority.

\subsection{Best Hyperparameters Across Super-Resolution Factors\label{sub:best_hp}}

We conducted experiments and hyperparameter search exclusively on the four pairs of SR factors mentioned above. In the Table 2 we present the best hyperparameters for each pair of factors. These values were selected by comparing the PITD of candidate models on the 2024 test period, from evaluation runs that covered different fractions of the test set, because the PITD is not computed during training. The selection is therefore not independent of the test data, and the comparison between candidates is approximate; the values in Table~2 should be read as indicative of the trend with $(S,T)$ rather than as optimal. A selection on a separate validation period, including a pixel-wise spread criterion, is left for future work.

\definecolor{inferno_1}{HTML}{fee08b} 
\definecolor{inferno_2}{HTML}{fdae61}
\definecolor{inferno_3}{HTML}{f46d43} 
\definecolor{inferno_4}{HTML}{d73027}

\begin{table}[h] 
\centering 
\begin{tabular}{ccccc}
\hline
\rowcolor{lightgray!50} Spatial SR factor $S$ & Temporal SR factor $T$ & Length $A_T$ & $\beta_{max}$ & Mass conservation function \\ 
\hline
1 & 3 & \cellcolor{inferno_1} 12 & \cellcolor{inferno_2} 1.5E-2 & \cellcolor{inferno_3} $f : x\rightarrow \sqrt{x}$, Threshold = 1E-2 \\
\hline
10 & 1 & \cellcolor{inferno_1} 10 & \cellcolor{inferno_1} 1E-2 & \cellcolor{inferno_3} $f : x\rightarrow \sqrt{x}$, Threshold = 1E-2 \\
\hline
10 & 3 & \cellcolor{inferno_3} 15 & \cellcolor{inferno_3} 2.E-2 & \cellcolor{inferno_1} $f : x\rightarrow x$, Threshold = 2E-2 \\
\hline
25 & 6 & \cellcolor{inferno_4} 18 & \cellcolor{inferno_4} 3.5E-2 & \cellcolor{inferno_1} $f : x\rightarrow x$, Threshold = 4E-2 \\
\hline

\end{tabular}
\caption{Optimal hyperparameter configurations as a function of the super-resolution factors. Three tuned hyperparameters are reported for each (S, T) setting. Warmer colors indicate higher values or fast-growing function in $[0, 1]$, highlighting scaling behavior with increasing SR factors}
\end{table}

Considering the attention time $A_T$, we observe that $A_T$ slightly increases as $T$ increases. This is consistent with expectations because when $T$ increases, the task is harder and the model needs more information to perform decent predictions. We also notice $A_T$ is not significantly greather than 12 which makes sense given that autocorrelation from precipitation is rarely significant beyond half a day.

From this table, it is apparent that increasing the SR factors, particularly $T$, leads to a growth in $\beta_{max}$. This is expected, because a larger $T$ (or $S$) increase the uncertainty and makes the task more difficult as the model must generate more pixels. Increasing $\beta_{max}$ allows the model to generate more diverse scenarios hopefully with some that better matches the ground truth.

Finally, this behavior is also observed for $F$, increasing the SR factors results in a function that grows slower (in $[0, 1]$) and limit the generation of extreme values. Indeed, the model tends to generate too many extreme precipitation events, it is helpful to apply this type of function $F$ to calibrate the model. We also notice the threshold we apply before the mass conservation function is increasing with the SR factors. Indeed, when the SR factors are high, the model is less likely to predict null predictions even though they represent most of the dataset, which is why we must increase this threshold.

\section{Conclusion}

To conclude, we successfully designed a versatile architecture capable of performing super-resolution at any arbitrary (space and time) granularity with minimal retuning. As expected from the bias--variance decomposition, the deterministic mean predictor remains best on pixel-wise and image metrics (MSE, MAE, 99th-percentile error, LSD, SSIM), while the generative model provides an ensemble with the lowest CRPS whose members restore the small-scale variability that the deterministic mean smooths out. The ensemble is still under-dispersed at the pixel scale; increasing its spread, through $\beta_{\max}$ and the threshold of $F$, is the main direction for improvement.

Nonetheless, several avenues for improvement can be envisioned. In particular, the architecture is not fully adaptable, since we allowed three hyperparameters to change. More importantly, the model requires one training instance per pair of super-resolution factors. Future work could aim to design a foundational architecture that shares weights and hyperparameters between factors to allow for seamless super-resolution across granularities.

\paragraph{Acknowledgments}
Central Computing Center at UNIL (DCSR). 

\paragraph{Funding Statement}
MD, FQ, and TB acknowledge support from the Swiss National Science Foundation (SNSF) under Grant No. 10001754 (``RobustSR'' project).

\paragraph{Competing Interests}
None.

\paragraph{Data Availability Statement}
The Coméphore dataset used in this project can be found on the French state website \url{https://www.data.gouv.fr/datasets/reanalyses-comephore/}.

\paragraph{Ethical Standards}
The research meets all ethical guidelines, including adherence to the legal requirements of the study country. 

\paragraph{Supplementary Material}
Supplementary material intended for publication has been provided with the submission, consisting of an appendix.

\paragraph{Author Contributions}
Conceptualization: T.B.; M.D.; F.Q.; M.V.; S.M. Methodology: T.B.; M.D.; F.Q.; M.V.; S.M.. Data curation: M.D. Experiments: M.D.; F.Q. Data visualisation: M.D. Writing original draft: T.B.; M.D.; F.Q; M.V.; S.M.. All authors approved the final submitted draft. 

\paragraph{Provenance Statement}
This article is part of the Climate Informatics 2026 proceedings and was accepted in Environmental Data Science on the basis of the Climate Informatics peer review process.



\bibliographystyle{unsrt}
\bibliography{bibliography}

\newpage
\begin{appendix}

\section*{Appendix}

\section{Deterministic Coarsening by Block Averaging}
\label{app:block_avg}

We formalize the deterministic HR to LR coarsening used in the perfect-model setting.
Let $S\in\mathbb{N}^*$ be the spatial SR factor and assume $S$ divides $H$ and $W$.
The LR grid has size $(H/S)\times(W/S)$.
Let $n$ and $m$ index LR pixels (i.e., spatial blocks): $n$ selects the $n$-th block in the vertical direction and $m$ the $m$-th block in the horizontal direction, with
$n\in\llbracket 0,\,H/S-1\rrbracket$ and $m\in\llbracket 0,\,W/S-1\rrbracket$.
The LR pixel $(n,m)$ corresponds to the $S\times S$ HR patch with HR indices
$(Sn,\ldots,Sn+S-1)\times(Sm,\ldots,Sm+S-1)$.

We define the spatial block-averaging operator
\begin{equation}
\mathcal{A}_S:\mathbb{R}_+^{H\times W}\to \mathbb{R}_+^{(H/S)\times(W/S)}
\end{equation}
by
\begin{equation}
(\mathcal{A}_S y)_{n,m}
\equiv
\frac{1}{S^2}\sum_{i=0}^{S-1}\sum_{j=0}^{S-1}
y_{Sn+i,\,Sm+j},
\qquad
n\in\llbracket 0,\,H/S-1\rrbracket,\;\;
m\in\llbracket 0,\,W/S-1\rrbracket.
\end{equation}

For temporal coarsening with factor $T\in\mathbb{N}^*$, we average over $T$ consecutive HR frames.
We define
\begin{equation}
\mathcal{A}_{S,T}:\big(\mathbb{R}_+^{H\times W}\big)^T \to \mathbb{R}_+^{(H/S)\times(W/S)}
\end{equation}
by
\begin{equation}
\mathcal{A}_{S,T}\!\left(y^{t,l},\ldots,y^{t+T-1,l}\right)
\equiv
\frac{1}{T}\sum_{k=0}^{T-1}\mathcal{A}_S\!\left(y^{t+k,l}\right),
\end{equation}
where $t$ indexes the start of the $T$-step temporal averaging block (and $l$ the tile).

Thus, in the main text the LR frame $x^{t,l}$ is obtained component-wise as
\begin{equation}
x^{t,l}_{n,m}
=
\frac{1}{T}\sum_{k=0}^{T-1}\frac{1}{S^2}\sum_{i=0}^{S-1}\sum_{j=0}^{S-1}
y^{t+k,l}_{Sn+i,\,Sm+j},
\qquad
n\in\llbracket 0,\,H/S-1\rrbracket,\;\;
m\in\llbracket 0,\,W/S-1\rrbracket.
\end{equation}

\section{Additional Modeling Details}
\label{details_diffusion_and_pictures}

\subsection{Schematic of the Deterministic U-Net Backbone\label{sub:U_Net}}

\begin{figure}[h!]
  \centering
  \includegraphics[width=\textwidth]{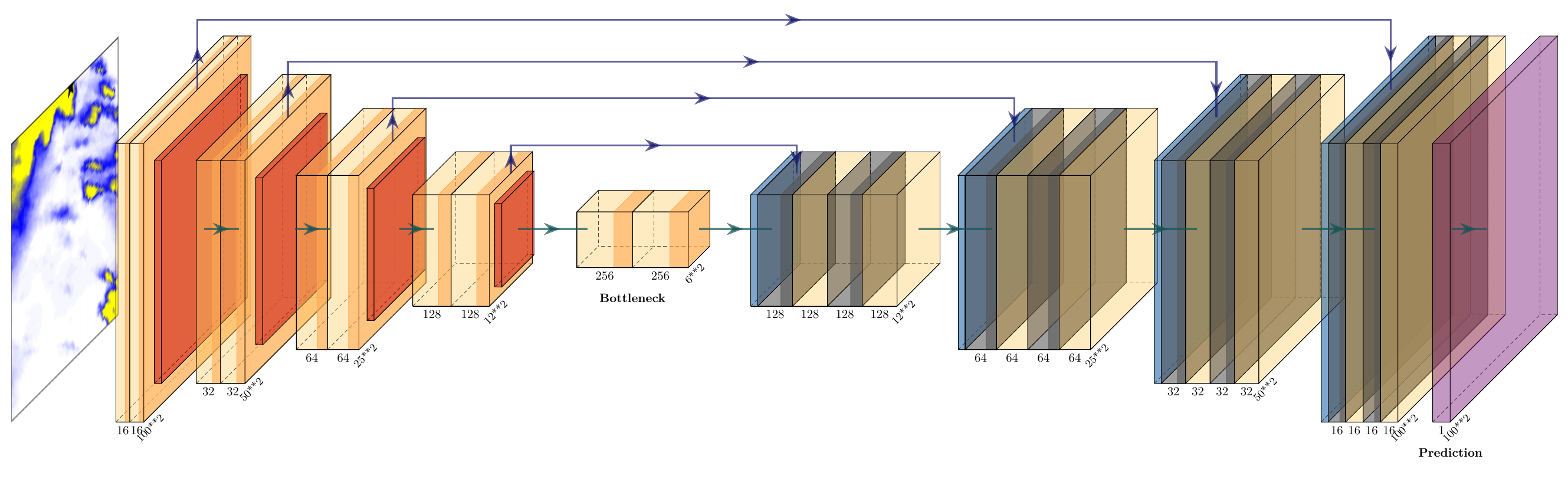}
  \caption{Proposed architecture for the deterministic U-Net. The encoder–decoder structure with convolutions, skip connections and attention module enables multiscale spatiotemporal feature extraction and reconstruction. Layer-wise channel shapes and counts are indicated}
  \label{figure_unet}
\end{figure}

In Figure \ref{figure_unet}, we describe the architecture of the deterministic U-Net, self attention modules are not represented but are used in each layer of the encoder. The diffusion U-Net follows the same architecture except for attention modules as explained in the main text.

\subsection{Forward process during diffusion}

We describe our forward process with the following equation, where $q$ is a gaussian distribution, $r^{t, l}(j)$ is the noised residual at step $j \in [1, J]$. The amount of noise we add at each step is controlled by the tuple $\beta \equiv (\beta_1, ..., \beta_J) \subset [0, 1]^J$. Throughout the rest of this section, we will also use $\alpha_j \equiv 1-\beta_j$ and $\sigma_j \equiv \sqrt{\beta_j}$. The forward process is given by :

\begin{equation}
    \forall j \in [1, J-1], \: q(r^{t, l}(j+1) \mid r^{t, l}(j)) \equiv \mathcal{N}(r^{t, l}(j+1) ; \sqrt{\alpha_j} \, r^{t, l}(j), \beta_j I)
\end{equation}

In fact, given the propriety of the Gaussian noise, the forward process comes down to sampling one Gaussian noise $\epsilon^{t, l} \sim \mathcal{N}(0, I)$ and adding it to the residuals with increasing intensity :

\begin{equation}
    r^{t, l}(j) = \sqrt{\bar{\alpha}_j} \, r^{t, l}(0) + \sqrt{1 - \bar{\alpha}_j} \, \epsilon^{t, l} \ \text{where}  \ \bar{\alpha}_j \equiv \prod_{s=1}^{j} (1-\beta_s)
    \label{forward_pass}
\end{equation}

\subsection{Velocity}
\label{velocity}

Given the noise and the initial frame, on can define the velocity as follows :

\begin{equation}
    v^{t, l}(j) = \sqrt{\bar{\alpha_j}}\times\epsilon^{t, l} - \sqrt{1-\bar{\alpha_j}}\times r^{t, l}(0)
    \label{velocity}
\end{equation}

The diffusion model returns $U_{dif}(r^{t, l}(j), BI(x^{t, l}), D^{t, l}, j)$ which is then compared, with a $L_2$ loss, to the true velocity. We also tried to train the model to predict the noise instead but we observed significantly worse performance. 

\subsection{Reverse process during diffusion\label{sub_reverse_process_diffusion}}
Concerning inference, we sample the residuals from a Gaussian noise $\hat{r}^{t, l}(J) \sim \mathcal{N}(0, I)$ and we denoise it iteratively by passing it to the U-Net concatenated with the same context as in training, especially the denoising step $j$. The model is trained to predict the velocity $\hat{v}^{t, l}(j)$ from which we can extract (thanks to Eq. \ref{forward_pass} and \ref{velocity})  the predicted noise as follow : 

\begin{equation}
    \hat{\epsilon}^{t, l}(j) = \sqrt{\bar{\alpha_j}}\times\hat{v}^{t, l}(j) + \sqrt{1-\bar{\alpha_j}}\times\hat{r}^{t, l}(j)
\end{equation}

Given that this predicted noise $\hat{\epsilon}$ is very unlikely to be perfectly spotted, we noise the prediction again to make it more realistic and keep the generative approach. Thus, by sampling $z \sim \mathcal{N}(0, I)$, the denoising process at step $j$ follows this equation : 

\begin{equation}
     \hat{r}^{t, l}(j-1) = \frac{1}{\sqrt{\alpha_j}} \left(  \hat{r}^{t, l}(j) - \frac{\beta_j}{\sqrt{1 - \bar{\alpha}_j}} \, \hat{\epsilon}^{t, l}(j) \right) + \sigma_j z
\end{equation}

This way, we recover denoised residuals up to $\hat{r}^{t, l}(0)$ and eventually reconstruct the frames with $\hat{y}^{t l} = D^{t', l} + \hat{r}^{t, l}(0)$.

\subsection{Algorithms describing the forward and reverse process}
\label{algorithm_forward_reverse}

{\centering
\begin{minipage}[t]{0.47\textwidth}
\begin{algorithm}[H]
\caption{Forward process}
\begin{algorithmic}[1]
\While{not converged}
\State Sample $x^{t, l}(L)$ and $y^{t, l} \in \text{Dataset}$
\State $D^{t, l} = \mathrm{MC}\!\left(F\!\left(U_{\mathrm{det}}([\mathrm{BI}(x^{t,l}(L)),M_l])\right),\,x^{t,l}\right)$
\State $r^{t, l}(0) = y^{t, l} - D^{t, l}$
\State Sample $j \sim \mathcal{U}[1, J]$ and $\epsilon \sim \mathcal{N}(0, I)$
\State $r^{t, l}(j) = \sqrt{\bar{\alpha}_j} \, r^{t, l}(0) + \sqrt{1 - \bar{\alpha}_j} \, \epsilon$
\State $\hat{v}^{t, l} = U_{dif}(r^{t, l}(j), BI(x^{t, l}), D^{t, l}, j)$
\State $v^{t, l}(j) = \sqrt{\bar{\alpha}_j} \cdot \epsilon - \sqrt{1 - \bar{\alpha}_j} \cdot r^{t, l}(0)$
\State Take gradient descent step on 

$\| \hat{v}^{t, l} - v^{t, l}(j) \|_2^2$
\EndWhile
\end{algorithmic}
\end{algorithm}
\end{minipage}
\hfill
\begin{minipage}[t]{0.52\textwidth}
\begin{algorithm}[H]
\caption{Reverse process}
\begin{algorithmic}[1]
\State Sample $x^{t, l}(L) \in \text{Dataset}$
\State $D^{t, l} = \mathrm{MC}\!\left(F\!\left(U_{\mathrm{det}}([\mathrm{BI}(x^{t,l}(L)),M_l])\right),\,x^{t,l}\right)$
\State Sample $\hat{r}^{t, l}(J) \sim \mathcal{N}(0, I)$ 
\For{$j = J$ down to $1$}
    \State $\hat{v}^{t, l} = U_{dif}(\hat{r}^{t, l}(j), BI(x^{t, l}), D^{t, l}, j)$
    \State $\hat{\epsilon}^{t, l} = \sqrt{\bar{\alpha}_j} \cdot \hat{v}^{t, l}(j) + \sqrt{1 - \bar{\alpha}_j} \cdot \hat{r}^{t, l}(j)$
    \State $z \sim \mathcal{N}(0, I)$ \textbf{if} $j \neq 1$ \textbf{else} $z = 0$
    \State $\hat{r}^{t, l}(j-1) =$
    
    $\frac{1}{\sqrt{\alpha_j}} \left( \hat{r}^{t, l}(j) - \frac{\beta_j}{\sqrt{1 - \bar{\alpha}_j}} \cdot \hat{\epsilon}^{t, l} \right) + \sigma_jz$
\EndFor
\State $\hat{y}^{t, l} = D^{t, l} + \hat{r}^{t, l}(0)$
\end{algorithmic}
\end{algorithm}
\end{minipage}}

\subsection{Overall architecture\label{sub:overall_architecture}}
\begin{figure}[h!]
  \centering
  \includegraphics[width=0.9\textwidth]{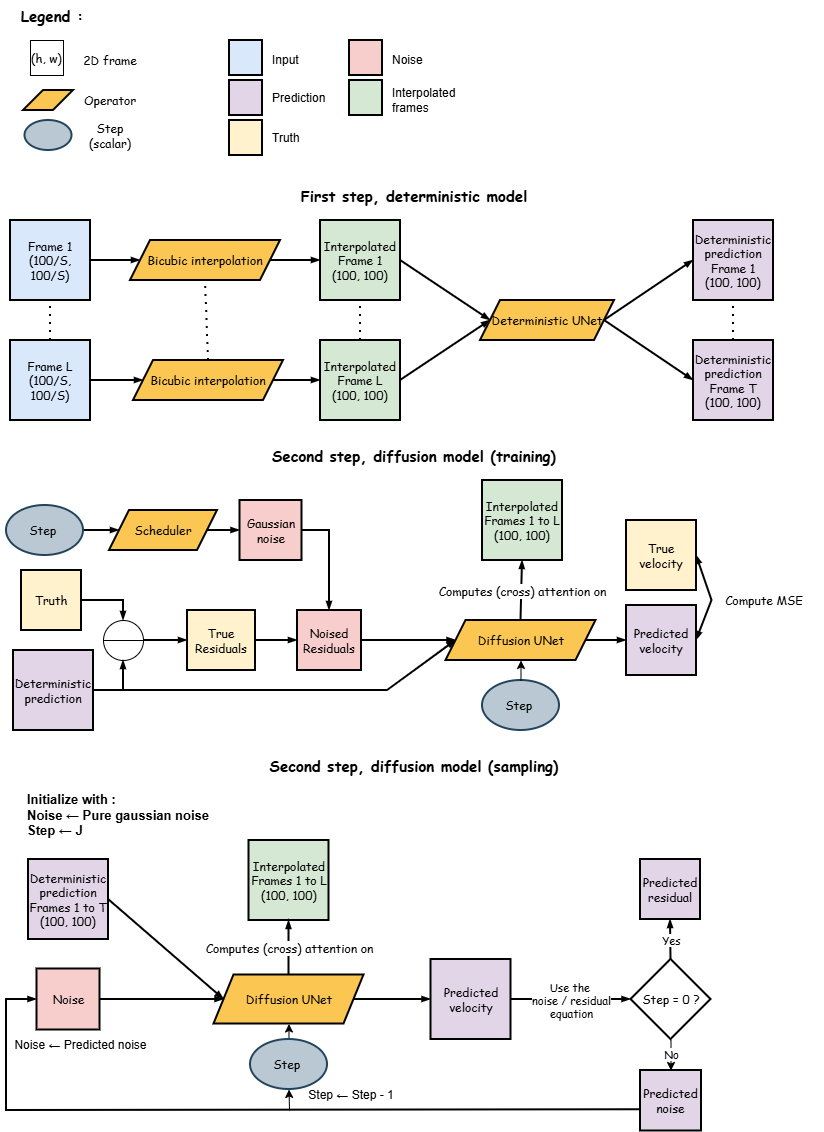}
  \caption{Schematic overview of the proposed super-resolution pipeline for precipitation. The workflow consists of an initial bicubic interpolation, followed by an average prediction module and a diffusion-based generative model for scenario generation. Separate pipelines are shown for training and inference, highlighting their distinct data flows and objectives}
  \label{figure_whole_architecture}
\end{figure}

\section{Geographical slicing}
\label{cross_val_slicing}

\begin{figure}[H]
  \centering
  \includegraphics[width=0.5\textwidth]{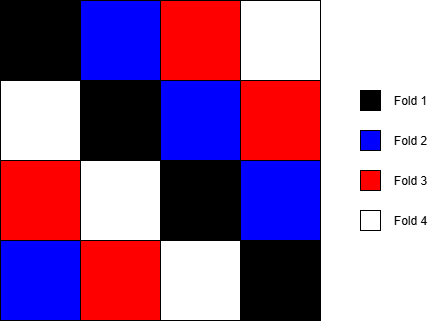}
  \caption{Spatial partitioning of the study domain into four cross-validation folds. Each fold contains one geographical zone per row and column, ensuring balanced spatial coverage}
\end{figure}

\section{Description of the metrics}\label{description_metrics}

\subsection{Metric computation protocol}\label{app:metric-computation}
We state precisely how each metric is computed for each class of model. Let the generative model produce $M$ ensemble members $\{\hat y^{(m)}\}_{m=1}^M$ and let $\bar y = \frac{1}{M}\sum_m \hat y^{(m)}$ be the ensemble mean.
\begin{itemize}
\item \textbf{Bias metrics (MSE, MAE, 99th~PE).} For deterministic models (including the interpolation baselines and EDSR), computed between the single prediction and the target. For the generative model, computed between the ensemble mean $\bar y$ and the target; we additionally report the mean per-member value $\frac{1}{M}\sum_m \mathrm{metric}(\hat y^{(m)}, y)$. For MSE, the mean per-member value exceeds the ensemble-mean value by exactly the mean squared spread of the members around their mean.
\item \textbf{Distributional metrics (CRPS, PITD).} Computed over the pooled distribution of all $M$ members, as described in the corresponding subsections; they assess the ensemble collectively rather than any single member.
\item \textbf{Spectral metric (LSD) and RAPSD.} Both start from the 2D spectrum of each frame, reduced to a 1D radial spectrum. The RAPSD averages the radial power spectra over frames (and members) and then compares them with the target (Figure~\ref{fig:rapsd}). The LSD compares the log radial spectra of prediction and target frame by frame, with a small constant $\epsilon$ inside the logarithm, and then averages over frames. In a frame where the target has no rain, its spectrum is zero, so any predicted rain gives a log-ratio of order $\log(1/\epsilon)$. More than half of the test frames are dry, and they dominate the LSD: in Table~\ref{tab:results_10_3} the deterministic model has the lowest LSD, but on the samples of Figure~\ref{fig:rapsd}, restricted to frames with rain, the generated members have a much lower LSD than the deterministic model (about $1.9$ against $3.0$ for light rain and $0.7$ against $1.2$ for heavy rain). Restricting the LSD to wavenumbers below the Nyquist limit does not change the ranking. We therefore use the RAPSD as the primary spectral diagnostic.
\item \textbf{Pixel-wise calibration.} The PITD (Appendix~\ref{description_metrics}) compares the distribution of values within a frame between the target and the pooled members; it does not test whether the ensemble covers the observation at each pixel. We therefore also report the rank of the observation among the $M$ members at each pixel (excluding pixels where the observation and all members are zero, ties broken at random) and the ratio of the ensemble spread $\sqrt{(1+1/M)\,\bar\sigma^2}$ to the root-mean-square error of the ensemble mean, which equals $1$ for a calibrated ensemble (Figure~\ref{fig:calibration}).
\item \textbf{Post-processing.} The deterministic U-Net applies the threshold ReLU $r_\alpha$ and the mass-conservation function $F$ to its output. The full architecture applies them to the deterministic mean $D$ and again to every member $D+\hat r$. The interpolation baselines are not post-processed.
\end{itemize}

\subsection{99th Percentile Error (99th PE)}
To evaluate the model’s capability in capturing extreme events, this metric considers the $99^{th}$ percentile of the data distribution and computes the absolute deviation between the predicted values and the ground truth. This metric is in $mm.h^{-1}$

\begin{equation}
99^{\text{th}}\text{PE} = |99^\text{th}(\text{predicted}) - 99^\text{th}(\text{ground truth})|
\end{equation}

\subsection{Log-spectral Distance (LSD)}
\begin{figure}[H]
\centering
\includegraphics[width=\linewidth]{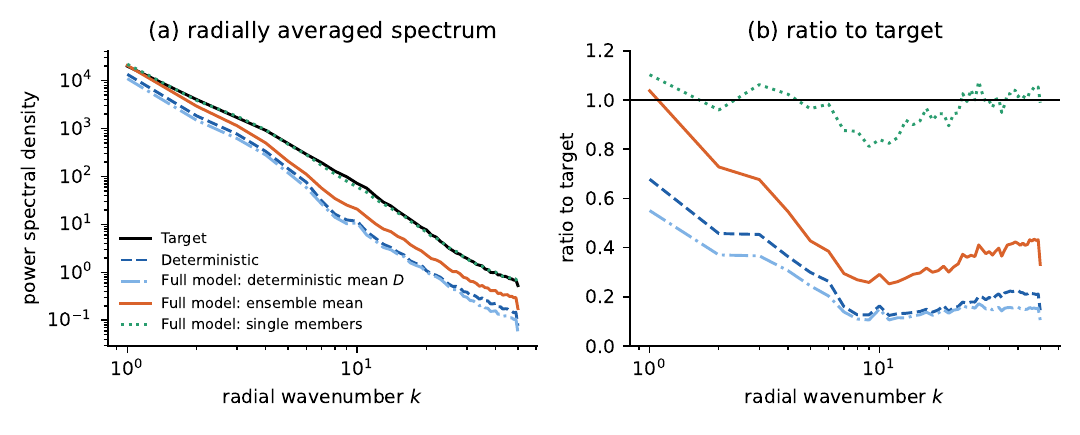}
\caption{Radially averaged power spectral density for $(S,T)=(10,3)$, computed on $576$ test frames ($192$ samples from four tiles, $8$ members). (a) Spectra; (b) ratio to the target spectrum. The single members match the target at all scales, while the deterministic prediction, the deterministic mean $D$ of the full model and the ensemble mean lose most of the small-scale power.}
\label{fig:rapsd}
\end{figure}

The log spectral distance was used to assess discrepancies in the frequency domain between predicted and reference images. For each image, a two-dimensional Fast Fourier Transform (2D FFT) was first computed, and the magnitude spectrum was extracted. The 2D spectrum was then mapped into a 1D spectrum followed by a logarithmic scaling of the spectral values. To transform the 2D frequencies $(n, m)$, the new frequency $f$ is computed as $f = \sqrt{n^2 + m^2}$, and its corresponding intensity is given by the arithmetic average of all the intensities whose mappings correspond to a new frequency $f$.
The LSD was finally obtained as the root mean square error between the log-spectra of the predicted and reference images. This metric has no units. To complement the scalar LSD, we also report the radially averaged power spectral density (RAPSD) for $(S,T)=(10,3)$. It is obtained by computing the 2D power spectrum per frame, reducing it to a 1D radial spectrum, and averaging over frames and over ensemble members; this per-frame-then-average procedure is standard in machine learning for weather. Figure~\ref{fig:rapsd} shows that the generated members reproduce the target spectrum at all scales, while the deterministic prediction and the ensemble mean lose most of the small-scale power.

\begin{figure}[H]
\centering
\includegraphics[width=\linewidth]{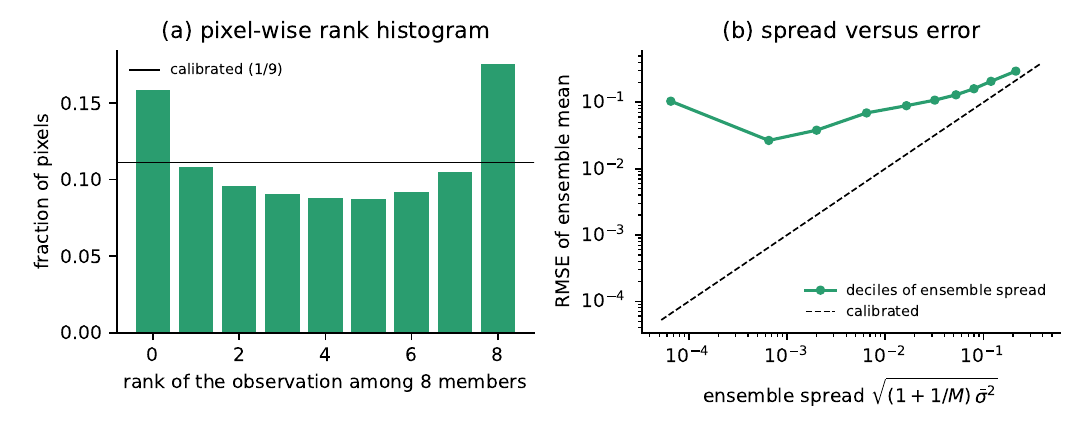}
\caption{Pixel-wise calibration of the $8$-member ensemble for $(S,T)=(10,3)$, on the same samples as Figure~\ref{fig:rapsd}. (a) Rank of the observation among the members: the U shape (outer bins $33\%$, against $22\%$ expected) shows an under-dispersed ensemble. (b) Root-mean-square error of the ensemble mean against the ensemble spread, by deciles of spread; a calibrated ensemble lies on the dashed line. The overall spread-to-error ratio is $0.58$. The leftmost point corresponds to pixels where all members predict almost no rain but rain is observed.}
\label{fig:calibration}
\end{figure}

\subsection{Earth-Moving Distance (EMD)}
The Earth-moving distance is a distribution metric. To compute it, we build the distribution of the prediction and the ground truth, then we compute the minimum cost of building one using the other. This metric is in $mm.h^{-1}$
More specifically, if $P$ and $T$ are respectively the predicted and target distribution, the EMD is computed as the supremum over the 1-Lipschitz continuous functions of the following quantity :
\begin{equation}
\text{EMD}(P,Q)=\sup \limits _{\|f\|_{L}\leq 1}\,\mathbb {E} _{x\sim P}[f(x)]-\mathbb {E} _{y\sim Q}[f(y)]\, 
\end{equation}

\subsection{Structural Similarity Index Measure (SSIM)}
The Structural Similarity Index (SSIM) is a widely used metric to quantify the perceptual similarity between two images. Unlike simple pixel-wise measures such as MSE, SSIM considers changes in structural information, luminance, and contrast, which better align with human visual perception. Given two image patches $x$ and $y$, the SSIM is defined as

\begin{equation}
\text{SSIM}(x, y) = \frac{(2 \mu_x \mu_y + C_1)(2 \sigma_{xy} + C_2)}{(\mu_x^2 + \mu_y^2 + C_1)(\sigma_x^2 + \sigma_y^2 + C_2)} 
\end{equation}

where $\mu_x$ and $\mu_y$ are the mean intensities of $x$ and $y$, $\sigma_x^2$ and $\sigma_y^2$ are the corresponding variances, $\sigma_{xy}$ is the covariance between $x$ and $y$, and $C_1, C_2$ are small constants to stabilize the division. The SSIM value ranges from $-1$ to $1$, with $1$ indicating perfect structural similarity, it is the only "score" of this article, which means the higher the better. To compute the SSIM metric for a video, we simply return the average of each SSIM between a predicted frame and a target frame. Finally, this metric has no unit.

\subsection{Probability Integral Transfort Deviation (PITD)}
\label{PITD}
The Probability Integral Transform Deviation (PITD) is a metric used to assess the calibration of probabilistic predictions. Given a predictive cumulative distribution function $F_i$ for the $i$-th target $y_i$, the probability integral transform is defined as

\begin{equation}
u_i = F_i(y_i)
\end{equation}

For a perfectly calibrated model, the values $u_i$ across all observations should follow a uniform distribution on the interval $[0,1]$ (see the proof in section \ref{proof_pitd}). The PITD quantifies the deviation of the empirical distribution of $u_i$ from this ideal uniform distribution. We compute it with the root-mean-square deviation from the expected uniform order statistics:

\begin{equation}
\text{PITD} = \sqrt{\frac{1}{N} \sum_{i=1}^{N} \left( u_i - \frac{i-0.5}{N} \right)^2 }
\end{equation}

where $N$ is the total number of observations. This metric has no unit. Lower PITD values indicate better calibration, while higher values reveal systematic biases in the predicted probabilities. In particular, two common cases of miscalibration can be observed. The first is the overdispersive case, where the model predominantly predicts extreme values, that is, either very low or very high. Conversely, the model can be underdispersed when it frequently produces values close to the median (this is often the case for interpolation baselines). 

In our case, we computed one Probability Integral Transform (PIT) $F_X(Y)$, where $X$ represents the predictions and $Y$ the target, for each observation $y$. We constructed the variable Y from the pixel values of the target image, thus, the distribution of Y simply corresponds to the distribution of the image's pixels. We did the same thing for $X$ but this time by considering all the pixels from the set of every generated scenarios. This decision has far-reaching consequences as it implies that we evaluate our model not on a single scenario (for example if we keep only the best one) but on the entire set of scenarios it generates. This approach is entirely desirable, as it accounts for the inherent uncertainty in the super-resolution task. Indeed, we are not interested in a model that, by sheer luck, produces a single scenario highly faithful to reality among a vast set of completely unrealistic ones. Rather, we prefer a model that generates a set of plausible scenarios, even if none of them is individually perfect. Because $X$ pools all pixels of a frame, this PIT tests whether the distribution of rain values in the generated frames matches that of the target frame, not whether the ensemble covers the observation at each pixel. In addition, the deviation is measured against the PIT of the members' own values, $F_X(X)$, rather than against a uniform distribution, which accounts for the large fraction of zeros. A low PITD therefore indicates realistic distributions of values; pixel-wise calibration is assessed separately (Appendix~\ref{app:metric-computation}, Figure~\ref{fig:calibration}). 

\begin{figure} [H]
    \centering
    \includegraphics[width=0.6\textwidth]{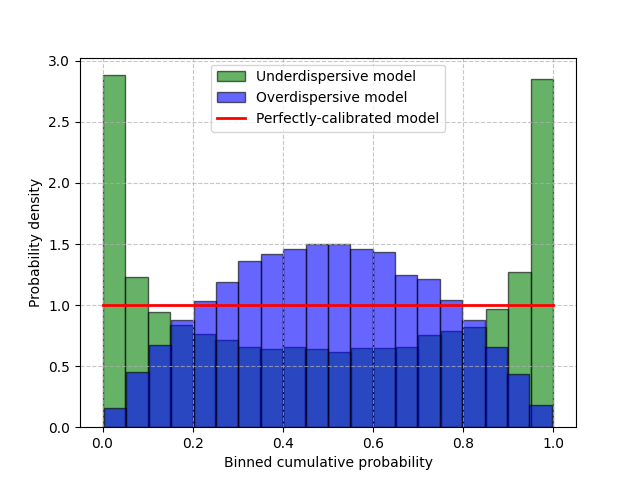}
    \caption{Illustration of Probability Integral Transform (PIT) behavior for under-dispersive and over-dispersive predictive distributions}
    \label{pitd_plot}
\end{figure}

With this modeling, a U-shaped PIT corresponds to an underdispersive model, whereas a bell-shaped PIT indicates an overdispersive one. The figure \ref{pitd_plot} compares the PIT of three differently calibrated models.

\subsection{Continuous Ranked Probability Score (CRPS)}

The Continuous Ranked Probability Score (CRPS) is a metric we use to evaluate the accuracy of probabilistic predictions. It measures the difference between the cumulative distribution function of the prediction and the target. Formally, for a forecast CDF $F$ and an observation $y$, the CRPS is defined as

\begin{equation}
    \text{CRPS}(F, y) = \int_{-\infty}^{\infty} \big( F(t) - \mathbf{1}\{t \ge y\} \big)^2 \, dt,
    \label{crps}
\end{equation}

where $\mathbf{1}\{t \ge y\}$ is the indicator function. 

In our case, the CRPS is computed with the same modeling as the PIT. Indeed, for a target frame $Y$, we compute a CRPS for each pixel then we average it to obtain a "frame" CRPS. To compute the CRPS over the pixel $(i, j)$, we set $y = Y_{i, j}$ and we build the variable $X$ such that its distribution corresponds to the distribution of the pixels $(i, j)$ from all the generated scenarios. Then we compute the CDF of $X$ and plug it into equation \ref{crps}. Again, this is coherent since we assess the model's predictions collectively rather than on a one-by-one basis. Anyway, its unit is $mm.h^{-1}$

We can show the CRPS generalizes the mean absolute error to probabilistic predictions. Indeed, for deterministic models, the CRPS coincides with the MAE as explained in Appendix \ref{crps_mae}.





\section{Proof of the Probability Integral Transform}
\label{proof_pitd}

Let $X$ be a continuous random variable with cumulative distribution function (CDF) $F_X$. Define a new random variable
\[
Y = F_X(X).
\]
We want to find the distribution of $Y$. The CDF of $Y$ is
\[
F_Y(y) = P(Y \leq y) = P(F_X(X) \leq y)
\]
Since $F_X$ is strictly increasing, we can apply its inverse:
\[
F_Y(y) = P(X \leq F_X^{-1}(y)) = F_X(F_X^{-1}(y)) = y
\]
Therefore, the CDF of $Y$ is
\[
F_Y(y) = y, \quad y \in [0,1]
\]
which is exactly the CDF of a uniform distribution on the interval $[0,1]$. 

This shows that if $X$ is a continuous random variable, then the transformed variable $Y = F_X(X)$ follows a uniform distribution on $[0,1]$.

\section{Proof of the equivalence between CRPS and MAE when the model is deterministic}
\label{crps_mae}

The Continuous Ranked Probability Score (CRPS) for a forecast CDF $F(t)$ and an observation $y$ is defined as:
\[
\text{CRPS}(F, y) = \int_{-\infty}^{\infty} \big(F(t) - \mathbf{1}\{t \ge y\}\big)^2 \, dt
\]
where $\mathbf{1}\{t \ge y\}$ is the indicator function.

For a deterministic prediction $y_0$, the forecast CDF becomes a step function:
\[
F(t) =
\begin{cases}
0 & \text{if } t < y_0, \\
1 & \text{if } t \ge y_0.
\end{cases}
\]

Plugging this into the CRPS definition:
\[
\text{CRPS}(y_0, y) = \int_{-\infty}^{\infty} \big( \mathbf{1}\{t \ge y_0\} - \mathbf{1}\{t \ge y\} \big)^2 \, dt
\]

The difference of the indicator functions is nonzero only on the interval between $y_0$ and $y$, and equals $\pm$1 there, and the square sets the value to 1. Therefore, the integral reduces to the length of this interval:
\[
\text{CRPS}(y_0, y) = |y_0 - y|
\]

Hence, for deterministic forecasts, the CRPS is exactly equal to the Mean Absolute Error:
\[
\text{CRPS}(y_0, y) = \text{MAE}(y_0, y)
\]
\end{appendix}

\end{document}